\documentclass[letterpaper, 10 pt, conference]{ieeeconf}
\IEEEoverridecommandlockouts

\usepackage{cite}
\usepackage{amsmath,amssymb,amsfonts}
\usepackage{tikz}
\usetikzlibrary{shapes.geometric, arrows.meta, positioning}
\usepackage{graphicx}
\usepackage{textcomp}
\usepackage{xcolor}
\usepackage{booktabs}
\usepackage{multirow}
\usepackage{url}
\usepackage{indentfirst}
\usepackage{tabularx}
\usepackage{placeins}
\usepackage[dvipsnames]{xcolor}
\usepackage{hyperref}
\definecolor{lightblue}{RGB}{135,206,250} 
\hypersetup{
    colorlinks=true,
    urlcolor=MidnightBlue,
    linkcolor=black,
    citecolor=lightblue
}
\usepackage{fontawesome5}
\usepackage[most]{tcolorbox}
\newtcbox{\linkpill}{
  on line,
  box align=base,
  colback=gray!10,
  colframe=gray!10,
  arc=1.5mm,
  boxrule=0pt,
  left=2pt,
  right=2pt,
  top=1pt,
  bottom=1pt
}
\definecolor{slsorange}{HTML}{D97706}
\definecolor{slsnavy}{HTML}{174A7E}

\def\BibTeX{{\rm B\kern-.05em{\sc i\kern-.025em b}\kern-.08em
    T\kern-.1667em\lower.7ex\hbox{E}\kern-.125emX}}

\begin{document}

\title{\LARGE \bf VerifLM: Physically Verified Image-Edit-Guided \\ Goal Pose Generation for Robotic Manipulation}

\author{Xuanyou Chen \and Jinyu Chen \and Glen Chou
\thanks{The authors are with the Georgia Institute of Technology, Atlanta, GA 30308, USA. \texttt{\{\href{mailto:xchen3019@gatech.edu}{xchen3019},\href{mailto:jchen3538@gatech.edu}{jchen3538},\href{mailto:chou@gatech.edu}{chou}\}@gatech.edu}}%
}
\maketitle

\begin{abstract}
Many manipulation goals are easier to visualize than to specify numerically through explicit geometric constraints. We present VerifLM, a framework that converts image-edited goal configurations into physically verified 6-DoF object poses for planning. Using an edited image, VerifLM recovers an initial relative pose between an active object and a passive object, then prompts a vision-language model to generate axis-wise sampling constraints and a task-specific validation program. Candidate poses are evaluated in a physics simulator, and the VLM selects the most task-consistent retained candidate. To operate without predefined object meshes or physical parameters, VerifLM reconstructs metric object geometry from RGB-D observations and constructs multiple hypothetical simulations with varying physics and geometry. Parallel-world validation evaluates candidates across geometry--physics variations and ranks them by the number of worlds in which they satisfy the validation program. We evaluate VerifLM on nine manipulation tasks. VerifLM achieves the highest task success on eight of nine tasks, with the largest performance gain on goals requiring precise orientation and contact relationships. 
\end{abstract}
\begin{center}
\vspace{-6.5pt}
{\linkpill{\href{https://veriflm.github.io/VerifLM-webpage/}{\textcolor{slsnavy}{\faGlobe\ \texttt{\textbf{Website}}}}}
\quad
\linkpill{\href{https://github.com/trustworthyrobotics/VerifLM}{\textcolor{slsnavy}{\faGithub\ \texttt{\textbf{Code}}}}}\quad
\linkpill{\href{https://youtu.be/HTQgtDVdVBo}{\textcolor{slsnavy}{\faYoutube\ \texttt{\textbf{Video}}}}}}
\vspace{-0.5pt}
\end{center}

\section{Introduction}
\begin{figure}[t]
    \centering
    \includegraphics[width=\columnwidth]{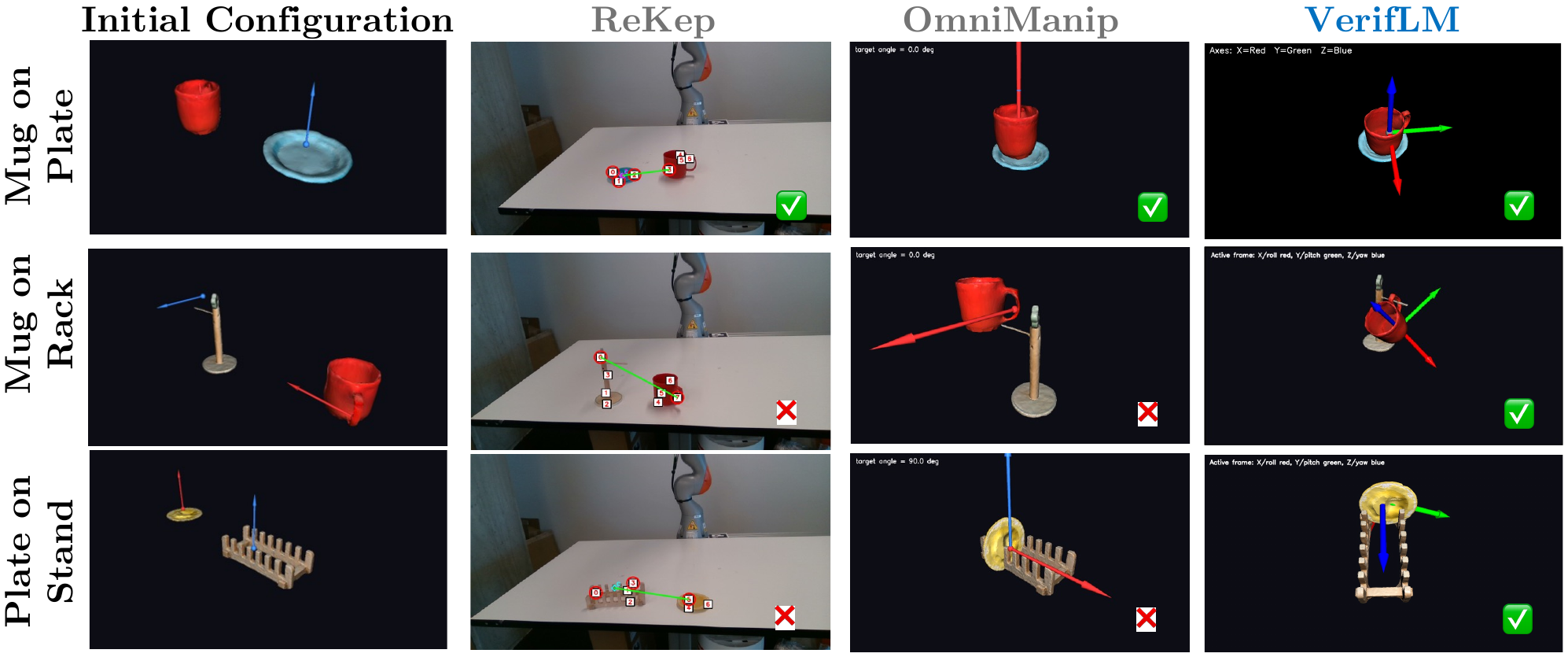}\vspace{-5pt}
    \caption{Illustration of ReKep and OmniManip constraints on representative tasks. Both methods perform well when object orientation is weakly constrained, as in Mug on Plate, where yaw does not affect success. However, they struggle when the intended goal pose requires a highly specific orientation or contact arrangement. In \textbf{Plate on Stand}, OmniManip recovers the desired orientation but not the position, causing penetration.\vspace{-15pt}}
    \label{fig:constraint-comparison}
\end{figure}

\begin{figure}[t]
    \centering
    \includegraphics[width=\columnwidth]{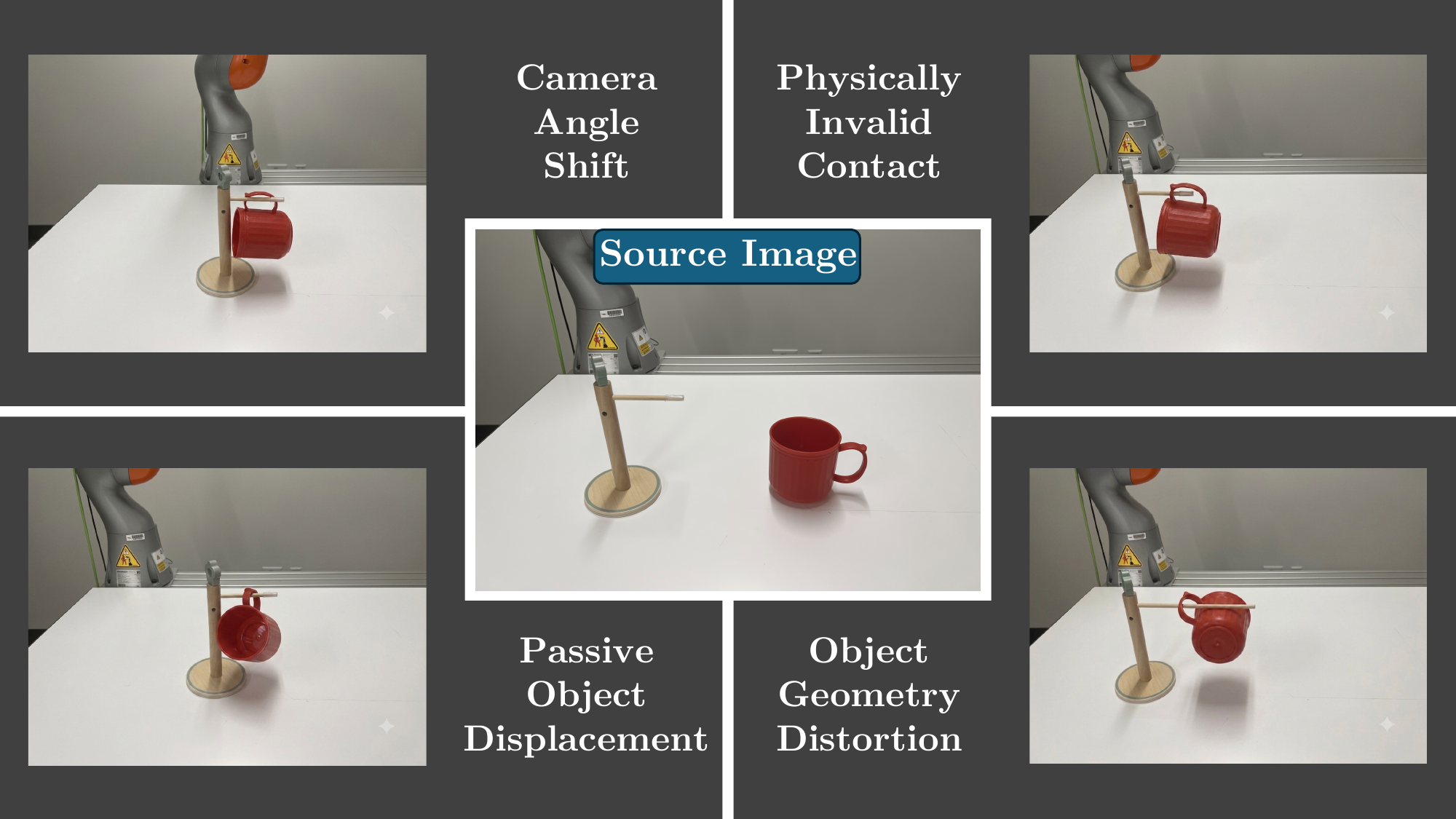}\vspace{-5pt}
    \caption{Common hallucination patterns in VLM-edited goal images.
    Given the same input scene, image editing may alter the camera viewpoint,
    distort object geometry, displace the passive object, or generate visually
    plausible but physically invalid contact.\vspace{-13pt}}
    \label{fig:hallucination-modes}
\end{figure}

A key challenge in robotic manipulation is inferring where and how an object should be placed from minimal human instructions, such as natural language. Consider inserting a plate into a plate stand, as shown in Fig.~\ref{fig:constraint-comparison}. Given an image of the scene, a human can readily visualize the desired configuration, yet specifying it \textit{numerically} requires identifying the appropriate slots and precisely positioning and orienting the plate between them. Thus, many manipulation goals are easier to visualize than to express as explicit poses. Vision-language models (VLMs), with their broad common-sense visual and semantic knowledge, offer a promising way to reason about such goals. However, VLMs typically output language, code, or images, while robot planners require numerical quantities such as object or end-effector poses. This raises a fundamental question: \emph{how can a VLM best infer a desired object configuration from natural language and vision to reliably extract a numerical manipulation goal?}
\looseness-1Existing methods interface VLMs with planners through executable programs and constraints \cite{codeaspolicies,owl_tamp}, dense 3D objectives \cite{voxposer,impact}, relational keypoints \cite{rekep}, or object-centric primitives \cite{omnimanip}. These representations integrate naturally with motion planners, but require the VLM to explicitly encode task geometry and can underconstrain precise 6-DoF goals. For example, ReKep~\cite{rekep} depends on visible, informative keypoints, while OmniManip~\cite{omnimanip} can leave rotation about an interaction axis unconstrained. As illustrated in Fig.~\ref{fig:constraint-comparison}, such pose ambiguity is acceptable for tasks like placing an upright mug on a plate, but problematic when success requires precise orientation or contact. Image editing offers an alternative: generative VLMs can directly visualize a language-specified goal in the observed scene \cite{instructpix2pix}, expressing detailed arrangements without explicitly formulating numerical geometric constraints. 
However, foundation models remain prone to spatial hallucinations \cite{kogilathota2026halp}, and VLM-edited images can contain geometrically inaccurate or physically implausible configurations \cite{phyedit}. Existing methods either recover goal poses assuming the edited image is accurate \cite{lamp}, or ask a VLM to revise erroneous edits before estimating the transformation \cite{goalvla}. As such, these methods rely heavily on purely visual verification, yet visually plausible configurations may be physically infeasible, as shown in Fig.~\ref{fig:hallucination-modes}. Many are also mesh-free and thus generalize to unseen objects, but depend on visual correspondence that can break under large rotations or viewpoint changes. Overall, it remains an open problem to infer goals that are precise, visually and physically plausible, and require minimal \textit{a priori} object geometry. 

To close these gaps, we propose \textbf{VerifLM}, a framework that uses image editing as an expressive interface for fine-grained object goals while treating the edited pose as an uncertain initialization for local, simulation-guided refinement. Our key observation is that image-editing errors are often structured: the predicted pose may be accurate along some translational and rotational axes but require substantial correction along others. Using a single RGB-D observation, VerifLM exploits this structure by using the VLM to determine \textit{where to search} more efficiently in pose space and \textit{which candidates are physically valid} by estimating scene geometry and physical properties to construct a digital twin online. To handle uncertainty in this reconstructed simulation, VerifLM evaluates candidates across an ensemble of plausible geometries and physical parameters and selects goals that remain valid across them. Our specific contributions are:
\begin{itemize}
\item We introduce \textit{VLM-guided goal refinement} that identifies uncertain pose dimensions, adapts their search ranges and directions, and generates task-specific physical validity criteria for candidate goals.
\item We enable \textit{physical verification without predefined object models} by reconstructing metric, simulation-ready geometry from a single RGB-D image and inferring multiple joint hypotheses over physical properties and geometry.
\item We introduce \textit{parallel-world ensemble validation}, which evaluates candidate poses across plausible geometry and physical-property hypotheses and ranks them by consistency with the simulated validation program.
\item We evaluate VerifLM on manipulation tasks requiring precise object poses and contacts, including settings with image-editing errors, large pose corrections, and uncertainty in online-reconstructed object models.
\end{itemize}

\section{Related Work}
\noindent\textbf{Foundation Models for Goal Specification. }
Foundation models have been used to map open-ended instructions into structured task representations, including executable programs and planner-compatible constraints~\cite{codeaspolicies,owl_tamp,pan2023data}, as well as keypoints, relational constraints, rewards, interaction primitives, dense value maps, and contact costs~\cite{moka,rekep,iker,omnimanip,voxposer,impact,qiu2026learning}. These interfaces expose foundation-model reasoning to conventional planning and control without directly predicting low-level actions. 
However, they require the desired behavior to be decomposed into an
explicit vocabulary of actions, points, directions, or geometric
relations. This can become under-parameterized when success depends
on a constrained goal configuration requiring precise
specification of translation, orientation, contact, or support. 
Rather than composing such primitives, our method uses an edited goal
image as a dense interface for specifying the complete relative object
configuration, then extracts, corrects, and physically validates the
corresponding $\mathrm{SE}(3)$ goal.

\looseness-1\noindent\textbf{Generated Visual Goals and Pose Inference. }
Generative visual models offer a dense interface for expressing desired scene changes. Image-editing models have been used to generate visual subgoals for learned controllers and visual chain-of-thought policies~\cite{instructpix2pix,susie,cotvla}, while recent methods lift generated visual goals into object transformations~\cite{goalvla,lamp} and 3D flows~\cite{dream2flow}. These methods exploit generative models' ability to express spatial relationships that are difficult to encode with sparse linguistic or geometric primitives. 
Related work also uses visual prompting to infer continuous spatial decisions. For instance, \cite{pivot} iteratively renders and refines candidate actions, while \cite{textguided6d} repeatedly queries a VLM for local translation or single-axis rotation updates. Although the latter targets language-consistent 6-DoF goal poses, its sequential steering can require many dependent VLM queries and does not physically validate visually plausible configurations. 
Our method instead uses a top-down approach: image editing first specifies the complete desired active--passive object configuration, from which we recover a relative $\mathrm{SE}(3)$ goal. The VLM then identifies uncertain pose components to constrain subsequent sampling, while simulation-based validation filters physically invalid candidates, refining the initial proposal into an executable goal.


\noindent\textbf{Digital Twins for Physically Grounded Goal Validation. }
Digital twins complement foundation-model reasoning with explicit geometric and physical prediction. Recent methods simulate candidate actions or strategies and present predicted outcomes to VLMs for evaluation and refinement~\cite{prompting_future,simpact,explicitworldmodel}. Related work uses simulation to validate object subgoals or transfer VLM-generated task specifications into policy optimization~\cite{adaptpnp,iker}, while reconstructed environments can also be augmented with inferred material properties for more realistic physics modeling and grasp-force reasoning~\cite{gaussianproperty}. 
These methods largely use simulation to evaluate proposals after they are generated. Our method also improves proposal generation itself. Instead of sampling broadly in an uninformed fashion, it diagnoses errors in the image-derived goal, identifies which translational and rotational components require correction, and restricts sampling to task-relevant regions of $\mathrm{SE}(3)$. The digital twin then filters candidates that violate geometric or physical constraints such as collision, penetration, support, and stability, before selecting the best task-consistent goal. 


\begin{figure*}[t]
    \centering
    \includegraphics[width=\textwidth, trim={0 18mm 0 25mm}, clip]{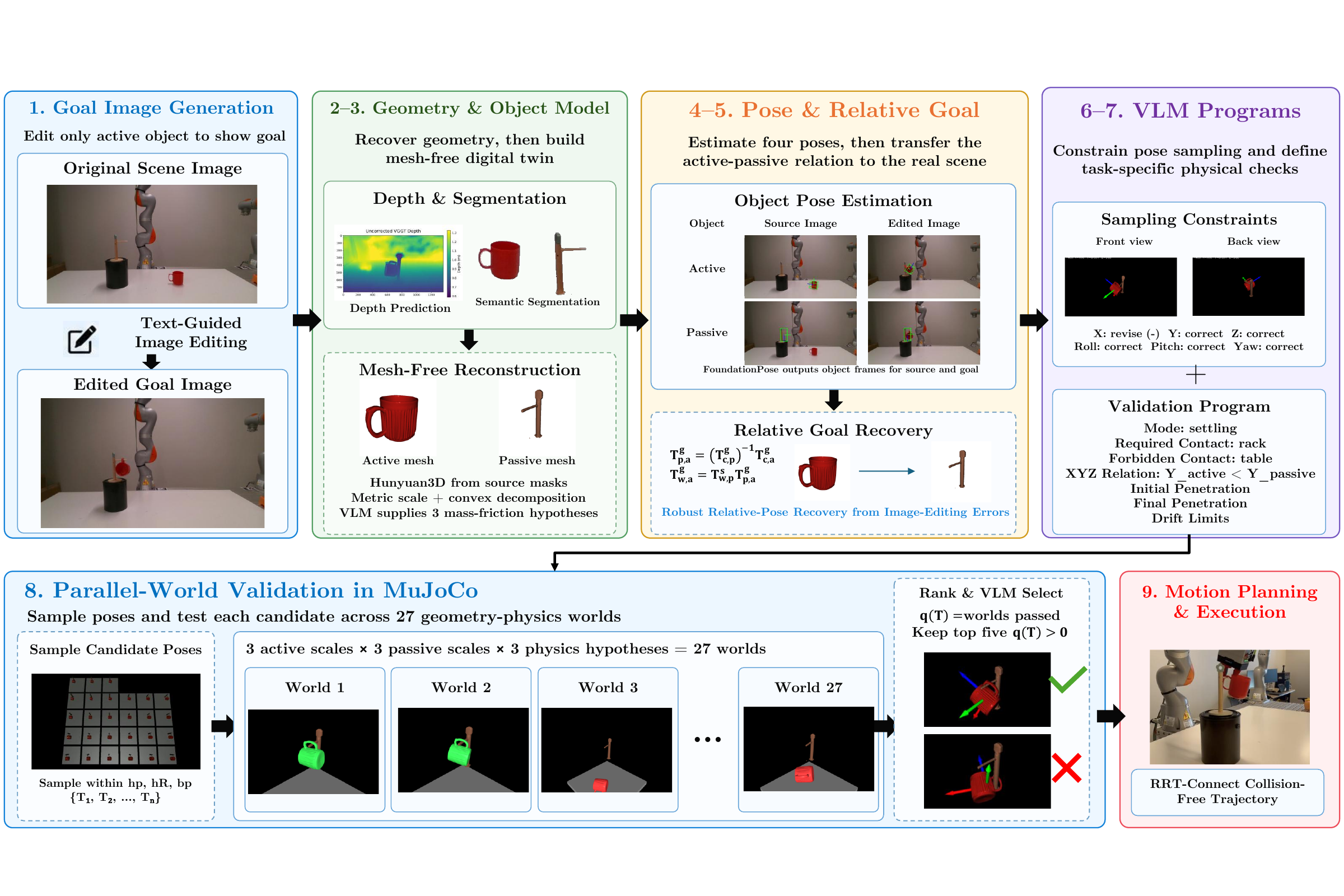}\vspace{-10pt}
    \caption{Overview of VerifLM, which converts an edited goal image into a physically verified object goal pose and executes a trajectory to realize it.\vspace{-33pt}}
    \label{fig:method_overview}\vspace{-13pt}
\end{figure*}

\section{Problem Formulation}
\label{sec:problem-formulation}
Given one source RGB-D observation $(I_s, D_s)$, a natural-language instruction $L$, an active object $O_a$, and a passive object $O_p$ that remains stationary throughout execution, our goal is to infer an instruction-consistent and physically valid goal pose for the active object $T_{w,a}^g\in SE(3)$ in the world frame. For instance, in the task of hanging a mug on a rack, the mug is the active object and the rack is a passive object. Given this goal pose, we subsequently find a feasible grasp and collision-free robot trajectory that realizes $T_{w,a}^g$. 

We consider two object-model settings. In the \textit{predefined-mesh} setting (\textbf{Problem 1}), object meshes and approximate physical parameters are known \textit{a priori}. In the \textit{mesh-free-input} setting (\textbf{Problem 2}), these are not known \textit{a priori}, and object geometry is reconstructed from the source RGB-D observation, while physical-parameter hypotheses are inferred from the source and edited RGB images, object identities, task instruction, and reconstructed dimensions. In both settings, we assume that the active and passive objects can be segmented and pose-estimated in both images and that the maximum possible sampling region around the recovered relative pose contains at least one task-consistent goal. In the mesh-free-input setting, we assume that the reconstruction preserves the task-defining contact geometry. For example, \texttt{Mug on Rack} requires an open mug handle and a suitably positioned rack branch, while \texttt{Plate on Stand} requires a disk-shaped object and the correct arrangement and clearance of rack slots.

\vspace{-6pt}
\section{Method}
\label{sec:method}
Given a source RGB-D observation and language instruction, VerifLM (Fig. \ref{fig:method_overview}) first generates an edited goal image and recovers the active object's pose relative to the passive object. A VLM then diagnoses uncertain pose dimensions and generates axis-wise sampling constraints and a task-specific validation program. We sample around the recovered proposal and physically validate candidates in MuJoCo. In the mesh-free-input mode (Problem 2), candidates are ranked across parallel digital twins with varying object scales and physical properties. Finally, a VLM selects the most task-consistent valid candidate for robot execution.

\subsection{Goal Image Generation}
\label{sec:goal-image-generation}
Given source image $I_s$, instruction $L$, active object $O_a$, and passive object $O_p$, we prompt an image-generative VLM to produce a goal image $I_g$ by moving or reorienting only $O_a$. Because the edit may contain geometric distortion, camera changes, incorrect contact, or passive-object displacement, its recovered pose is treated only as an initial proposal.

\subsection{Depth and Object Segmentation}
\label{sec:depth-and-segmentation}
Since the edited goal image has no associated metric depth, we jointly estimate the depths of the source and edited goal images using VGGT~\cite{vggt}, obtaining $\tilde{D}_s$ and $\tilde{D}_g$. We use the source RGB-D measurement $D_s$ to estimate correction scale $s$ and offset $t$ for the VGGT-predicted depth by solving 
\begin{equation}
(s,t)=\arg \min_{s,t}\sum_{\mathbf{u}\in\Omega}[D_s(\mathbf{u})-(s\tilde{D}_s(\mathbf{u})+t)]^2,
\end{equation}
where $\mathbf{u}$ indexes image pixels and $\Omega$ denotes the set of pixels with valid source depth measurements. We then apply the fitted correction to the edited-image depth
$D_g=s\tilde{D}_g+t$. 
We next localize the active and passive objects in both images using GroundingDINO\cite{groundingdino} followed by SAM2\cite{sam2} segmentation.

\subsection{Object Model and Pose Estimation}
\label{sec:pose-estimation}
\looseness-1In \textit{predefined-mesh} mode, we use object meshes obtained with Apple Reality Composer and manually specified physical parameters. In \textit{mesh-free-input} mode, we instead reconstruct meshes from the source RGB-D observation using Hunyuan3D~\cite{hunyuan3d21} on segmented object crops. To recover metric scale, we back-project masked depth pixels into a source-camera point cloud and remove disconnected background points based on depth discontinuities. We center the remaining points, compute their principal axes, and estimate robust extents from the 1st--99th percentile range along each axis. Let $\ell_{\mathrm{obs}}$ be the largest observed extent and $\ell_{\mathrm{mesh}}$ the largest bounding-box extent of the generated mesh. We then uniformly scale the mesh by $s_{\mathrm{mesh}}=\frac{\ell_{\mathrm{obs}}}{\ell_{\mathrm{mesh}}}$. 
As only visible surfaces are observed, this scale is estimated, and parallel-world validation accounts for estimation error. Finally, meshes are convex-decomposed to obtain MuJoCo collision geometry.

\looseness-1Since mass and friction cannot be recovered directly from a single RGB-D observation, we prompt a VLM with the instruction, source and edited images, object names, and reconstructed dimensions. The VLM jointly predicts three hypotheses for the active and passive objects, where each hypothesis specifies a mass and sliding-friction coefficient for both objects. Joint estimation encourages physically consistent parameters within each active--passive pair. 
Given the object masks, either predefined or reconstructed object meshes, camera intrinsics, and the corresponding depth maps, FoundationPose \cite{foundationpose} estimates four active/passive object poses 
$T_{c,a}^{s}$, $T_{c,p}^{s}$, $T_{c,a}^{g}$, and $T_{c,p}^{g}$,
where $c$ denotes the camera frame and superscripts $s$ and $g$ denote the source and goal images, respectively. 

\subsection{Relative Goal Pose Recovery}
\label{sec:relative-goal-pose-recovery}
\looseness-1To reduce sensitivity to global editing errors, such as camera or scene displacement, we use the edited image only to recover the active object's pose relative to the passive object:
\begin{equation}
    T_{p,a}^g=\left(T_{c,p}^{g}\right)^{-1}T_{c,a}^{g}.
\end{equation}
We apply this relative configuration to the passive object's world-frame pose to yield the active object's world-frame goal pose. Here, $T_{c,w}$ maps world-frame coordinates to camera-frame coordinates.
\begin{equation}
    T_{w,a}^{g}
    =(T_{c,w})^{-1}T_{c,p}^{s}T_{p,a}^g.
\end{equation}

\subsection{VLM-Generated Sampling Constraints}
\label{sec:sampling-constraints}
Because $T_{p,a}^g$ can still reflect edit-time hallucination, it is treated as a
\textit{proposal} rather than a final target. Importantly, the proposed active-object pose is often
accurate along several translational and rotational axes and requires
corrections along only certain others. Therefore, rather
than sampling uniformly around the proposal, we ask the VLM to
identify which axes are correct or close to the desired configuration, and which axes require revision. Translation and rotation are diagnosed separately. 
For translation, we render the proposed configuration from multiple
viewpoints with the passive-object coordinate frame visualized. The VLM
assesses, for each axis of the passive-object frame, how strongly the
proposal should be perturbed and whether the correction should preferentially
occur in the positive or negative direction. We denote the resulting
axis-wise translational half-ranges and directional preferences by
$\mathbf{h}_p = [h_{p,x},h_{p,y},h_{p,z}]$ and  
    $\mathbf{b}_p = [b_{p,x},b_{p,y},b_{p,z}]$, 
where the translational half-ranges are obtained by multiplying the VLM-selected scale factors by the active object's characteristic size $L_a$ (e.g., longest dimension). This scaling allows the same
reasoning procedure to adapt across objects of different physical sizes.

\looseness-1We similarly reason about orientation by rendering the proposal with the
active object's local frame. The VLM determines which rotational
degrees of freedom are already consistent with the desired configuration
and which require broader refinement, yielding axis-wise rotational search
scales
$\mathbf{h}_R = [h_{R,\mathrm{roll}},
                    h_{R,\mathrm{pitch}},
                    h_{R,\mathrm{yaw}}]$. 
Larger search scales are assigned to pose components judged to require
greater correction, while components already consistent with the task are
searched only locally. 
Together, these quantities define the VLM-conditioned sampling constraint
$\mathcal{S} = (\mathbf{h}_p,\mathbf{h}_R,\mathbf{b}_p)$, 
which specifies a task-adaptive neighborhood of the image-derived proposal
in $SE(3)$. Section~\ref{sec:mujoco-sample-and-validation} describes how samples are drawn from
this region and physically validated in the digital twin.

\subsection{VLM-Generated Validation Program}
\label{sec:validation-program}
The sampling constraints $\mathcal{S}$ determine where to sample candidate poses. Each sample is evaluated by a task-specific validation program $\mathcal{V}$ in the digital twin and retained or discarded accordingly. Given the instruction and rendered scene observations, the VLM predicts
\begin{equation}
\mathcal{V}
=
(
m,\,
\mathcal{C}_{\mathrm{req}},\,
\mathcal{C}_{\mathrm{forb}},\,
\mathcal{R}_{xyz},\,
\boldsymbol{\epsilon}_{p},\,
\boldsymbol{\epsilon}_{R}
).
\end{equation}
Here, $m\in\{\texttt{settling},\texttt{geometric\_only}\}$ is the
validation mode, $\mathcal{C}_{\mathrm{req}}$ and
$\mathcal{C}_{\mathrm{forb}}$ specify required and forbidden contacts,
and $\mathcal{R}_{xyz}$ specifies object-center relations. The vectors
$\boldsymbol{\epsilon}_p$ and $\boldsymbol{\epsilon}_R$ define the
allowable per-axis translational and rotational drift between the
proposed and settled poses. For settling tasks, the VLM specifies per-axis tolerances on the
translational and rotational change permitted during physical relaxation.
Candidate poses are then simulated under gravity and evaluated using the
resulting settled configuration. For geometric-only tasks, applicable
criteria are instead evaluated directly at the sampled pose without forward
simulation. We additionally allow limited initial interpenetration during
settling so that the simulator may resolve small geometric inconsistencies
into physically feasible configurations, while requiring the final settled
state to satisfy a substantially tighter penetration criterion.
Table~\ref{tab:validation-timing} summarizes the validation criteria and
their evaluation timing. 

\begin{table}[t]
\centering
\caption{Validation criteria, evaluation timing, and whether manually configured or VLM-provided.\vspace{-5pt}}
\label{tab:validation-timing}
\resizebox{\columnwidth}{!}{
\begin{tabular}{lccccc}
\toprule
Criterion
& Initial pose
& Every settling step
& Final state
& \texttt{geometric\_only}
& Source \\
\midrule
Initial penetration
& Yes & No & No & No
& Configured \\
Final penetration
& No & No & Yes & Yes
& Configured \\
Forbidden contacts
& Yes & Yes, immediate rejection & Yes & Yes
& VLM \\
Required contacts
& No & No & Yes & Yes
& VLM \\
Position drift
& No & No & Yes & No
& VLM \\
Orientation drift
& No & No & Yes & No
& VLM \\
XYZ relations
& No & No & Yes & Yes
& VLM \\
\bottomrule
\end{tabular}
}\vspace{-18pt}
\end{table}

\subsection{MuJoCo Sampling and Validation}
\label{sec:mujoco-sample-and-validation}
Given sampling constraints $\mathcal{S}$, we generate candidate poses uniformly around the image-derived relative pose $T_{p,a}^g$. Let $(p_0,R_0)$ denote its translation and rotation. 
For candidate $i$, 
\begin{align} 
p_i &= p_0 + \delta p_i, & \delta p_{i,j} &\sim U(-h_{p,j},h_{p,j}), \\ 
R_i &= R_0 \Delta R(\delta\boldsymbol{\theta}_i), & \delta\theta_{i,j} &\sim U(-h_{R,j},h_{R,j}), 
\end{align} \looseness-1where translational perturbations are expressed in the \textit{passive object frame}, while $\Delta R$ applies rotational perturbations about the \textit{active object's local frame}. If $b_{p,j}$ specifies a positive or negative directional bias, the corresponding translational interval is restricted to $[0,h_{p,j}]$ or $[-h_{p,j},0]$, respectively. The unperturbed proposal is evaluated before drawing perturbed samples. 
Each sampled relative pose is transformed into the world frame using the measured passive-object pose, \begin{equation} T_{w,a}^{(i)} = T_{w,p}^{s}T_{p,a}^{(i)}. \end{equation} In predefined-mesh mode, the candidate is instantiated in a MuJoCo~\cite{mujoco} digital twin containing the table, robot geometry, passive object, and object physical properties. The candidate is then evaluated according to the validation program $\mathcal{V}$ in Sec.~\ref{sec:validation-program}. Only candidates that pass all criteria are kept. Sampling continues until $N_c$ valid candidates are obtained or the sampling budget of $N_s$ proposals is exhausted. The candidates are represented by the active object's settled pose.

In mesh-free-input mode, reconstructed geometry and predicted physics are only estimated. Thus, in our setting, we evaluate each candidate in an ensemble of $N_w=27$ collision-isolated parallel worlds within one simulation environment. The worlds enumerate three scale factors $\{0.9,1.0,1.1\}$ for each of the active and passive objects and the three jointly predicted physics hypotheses, i.e., $N_w=3\ \text{active scales}\times3\ \text{passive scales}\times3\ \text{physics hypotheses}$. 
Each world applies the same validation program $\mathcal{V}$. We assign candidate $T$ the robustness score
\begin{equation}
q(T)=\textstyle\sum_{w=1}^{N_w}\mathbb{I}\!\left[\mathcal{V}(T;w)=1\right],
\end{equation}
where $\mathcal{V}(T;w)=1$ indicates that candidate $T$ passes validation in world $w$. After exhausting the sampling budget, we retain the $N_c$ highest-scoring candidates with $q(T)>0$, except that geometric-only validation always preserves the initial proposal. For each retained proposal, we examine the settled poses from the worlds in which it successfully satisfies $\mathcal{V}$. Each settled pose is transferred to the nominal-scale geometry, with both object scales set to 1.0, and we retain the first pose whose maximum penetration depth does not exceed a specified tolerance. If no successful world yields a nominal-scale pose within this tolerance, the proposal is discarded. Finally, the retained candidates are rendered with the active object's coordinate axes visualized, and the VLM selects the candidate that best satisfies the task instruction.

\subsection{Grasp and Motion Planning}
\label{sec:grasp-and-motion-planning}
Once $T_{w,a}^g$ has been selected, it can be paired with any motion planner to generate a collision-free trajectory to achieve it. The current system uses AnyGrasp\cite{anygrasp} to generate candidate grasps, filtered for mask consistency, collision-free start and goal configurations, and IK feasibility. RRTConnect\cite{rrtconnect} is then used to plan a collision-free trajectory, time-parametrized according to a maximum joint velocity.

\subsection{Multi-stage Task Extension}
\label{sec:grasp-planning}
For multi-stage tasks, we repeat the pipeline for each configured active--passive pair using a new RGB-D observation. The current pair is re-estimated from the updated scene, while previously manipulated objects outside the pair are represented as obstacles using their selected goal poses. This limits the accumulation of pose errors across stages.

\begin{table}[t]
\centering
\caption{Manipulation benchmarks and task-level success criteria.\vspace{-5pt}}
\label{tab:benchmarks}
\small
\begin{tabularx}{\columnwidth}{l X}
\toprule
Single-Stage Task & Success criterion \\
\midrule
\texttt{Mug on Rack}
& Mug hangs from the rack by its handle. \\

\texttt{Mug on Plate}
& Mug is upright near the plate center. \\

\texttt{Mug on Stand}
& Mug is placed on the plate stand inverted, covering at least two notches. \\

\texttt{Plate on Stand}
& Plate is securely supported in a pair of opposing slots, either directly corresponding or offset by one slot. \\

\texttt{Pour Tea}
& Teapot reaches a valid pouring pose with its spout over the mug. \\

\texttt{Lid on Teapot}
& Lid is correctly seated on the teapot. \\

\texttt{Stack Cup}
& One cup is placed inside the other. \\\\

\toprule
Multi-Stage Task & Success criterion \\
\midrule

\texttt{Plates on Stand}
& Both plates are securely supported in pairs of opposing slots, either directly corresponding or offset by one slot. \\

\texttt{Stack Cups}
& All cups are successfully stacked in the specified order. \\

\bottomrule
\end{tabularx}\vspace{-15pt}
\end{table}

\vspace{-5pt}
\section{Experimental Results}
\label{sec:experiments}

We aim to answer the following research questions:
\begin{enumerate}
    \item Can VerifLM reliably recover and execute instruction-consistent,
    physically valid goal poses across a diverse set of manipulation tasks?
    \item How do the VLM-generated sampling constraints $\mathcal{S}$, validation
program $\mathcal{V}$, and parallel-world validation contribute to task success?
\end{enumerate}

\subsection{Experimental Setup}
\label{sec:exp-setup}
We evaluate VerifLM on a KUKA iiwa 7 R800 manipulator with an Intel RealSense D415 camera for RGB-D input. We use Gemini~3.1 Flash Image for goal-image editing and GPT-5.6 for generating the sampling constraints, validation program, joint physical-parameter hypotheses, and final candidate selection. Our benchmark suite contains nine household manipulation tasks spanning placement, hanging, insertion,
pouring, and stacking, with two multi-stage tasks. Table~\ref{tab:benchmarks} summarizes the
tasks and their success criteria.

Each method is evaluated for 24 trials per task with randomized object placements. The 24 trials are
divided into three groups of eight, and we report both the total number of
successful trials and the mean $\pm$ standard deviation of the success rate
across the three groups. We compare against ReKep \cite{rekep}, OmniManip \cite{omnimanip}, and Goal-VLA \cite{goalvla}. ReKep uses
automatically generated keypoints. Table~\ref{tab:main-results} separately reports VerifLM
with predefined object models and with reconstructed object models. 
In our implementation, the VLM expresses each search scale (Sec. \ref{sec:sampling-constraints}) using three
ordered levels---\emph{local}, \emph{moderate}, and \emph{broad}---which
are mapped to increasing perturbation ranges. We use translational
half-ranges of $0.25L_a$, $0.75L_a$, and $1.5L_a$, and rotational
half-ranges of $20^\circ$, $60^\circ$, and $180^\circ$, respectively.
In Sec. \ref{sec:validation-program}, the VLM's \emph{strict}, \emph{moderate}, and
\emph{permissive} drift levels correspond to translational tolerances of
$0.25L_a$, $0.5L_a$, and $0.75L_a$ and rotational tolerances of
$20^\circ$, $40^\circ$, and $60^\circ$, respectively. We use initial and
final penetration thresholds of $10$\,mm and $0.6$\,mm.

\subsection{Quantitative Results}
\label{sec:quantitative-results}
Table~\ref{tab:main-results} reports task success for all nine benchmarks.
For each entry, we report the total number of successful trials out of 24
and, in parentheses, the mean $\pm$ standard deviation of success rate
across three groups of eight trials. Task-planning success requires the evaluated object configuration to satisfy the task-level success criterion in Table~\ref{tab:benchmarks} and remain within the final penetration tolerance of $\mathcal{V}$. Execution success additionally requires the physical robot to successfully move the object in the real world to a task-consistent configuration.

VerifLM achieves the highest task-planning success on eight of the nine tasks, with \texttt{Pour Tea} as the only exception. With predefined meshes, VerifLM achieves 160/216 (74.1\%) task-planning success and 155/216 (71.8\%) execution success, compared with 102/216 (47.2\%) for the strongest baseline overall, Goal-VLA. Relative to the best baseline for each task, the largest planning gains occur on \texttt{Plates on Stand} (+33.3 percentage points), \texttt{Plate on Stand} (+29.2 percentage points), and both \texttt{Stack Cup} and \texttt{Stack Cups} (+25.0 percentage points). The smaller gain on Mug on Plate (+8.3 points), where substantial yaw ambiguity is acceptable, suggests that keypoints and object-centric primitives can be sufficient for tasks whose goal poses are loosely constrained, whereas VerifLM is most beneficial when success requires a precise and hard-to-describe orientation or contact arrangement. 
Without predefined meshes or physical parameters, VerifLM still achieves 146/216 (67.6\%) task-planning success and 130/216 (60.2\%) execution success, but trails the predefined-mesh variant by 6.5 and 11.6 percentage points, respectively. This gap reflects the sensitivity of downstream validation and execution to reconstructed geometry. In cup stacking, for example, an oversized estimated active cup paired with an undersized estimated passive cup can prevent the cups from being stacked altogether, significantly reducing task-planning success. Similarly, a small geometric or pose error in Lid on Teapot can determine whether the lid settles stably or slides off. Parallel-world validation mitigates these errors by evaluating candidates across multiple scales and physics hypotheses, but cannot fully compensate for the discrepancies in geometry.

\begin{table*}[t]
\centering
\caption{Task-planning and execution success across nine benchmarks.}
\label{tab:main-results}
\resizebox{\textwidth}{!}{
\begin{tabular}{lccccccc}
\toprule
Task
& ReKep~\cite{rekep}
& OmniManip~\cite{omnimanip}
& Goal-VLA~\cite{goalvla}
& \multicolumn{2}{c}{Ours (with predefined meshes)}
& \multicolumn{2}{c}{Ours (mesh-free-input)} \\
\cmidrule(lr){5-6}
\cmidrule(lr){7-8}
& Task Planning
& Task Planning
& Task Planning
& Task Planning
& Execution
& Task Planning
& Execution \\
\midrule

\multicolumn{8}{l}{\textit{Single-stage}} \\

Mug on Rack
& 4/24 (16.67\%$\pm$11.79\%)
& 6/24 (25.00\%$\pm$10.21\%)
& 11/24 (45.83\%$\pm$5.89\%)
& 16/24 (66.67\%$\pm$15.59\%)
& 16/24 (66.67\%$\pm$15.59\%)
& \textbf{17/24 (70.83\%$\pm$11.79\%)}
& 14/24 (58.33\%$\pm$11.79\%) \\

Mug on Plate
& 16/24 (66.67\%$\pm$15.59\%)
& 19/24 (79.17\%$\pm$5.89\%)
& 20/24 (83.33\%$\pm$11.79\%)
& \textbf{22/24 (91.67\%$\pm$11.79\%)}
& \textbf{22/24 (91.67\%$\pm$11.79\%)}
& 20/24 (83.33\%$\pm$15.59\%)
& 20/24 (83.33\%$\pm$15.59\%) \\

Mug on Stand
& 10/24 (41.67\%$\pm$11.79\%)
& 7/24 (29.17\%$\pm$11.79\%)
& 15/24 (62.50\%$\pm$17.68\%)
& \textbf{20/24 (83.33\%$\pm$15.59\%)}
& \textbf{20/24 (83.33\%$\pm$15.59\%)}
& 18/24 (75.00\%$\pm$17.68\%)
& 17/24 (70.83\%$\pm$15.59\%) \\

Plate on Stand
& 0/24 (0.00\%$\pm$0.00\%)
& 6/24 (25.00\%$\pm$10.21\%)
& 10/24 (41.67\%$\pm$5.89\%)
& \textbf{17/24 (70.83\%$\pm$11.79\%)}
& 16/24 (66.67\%$\pm$15.59\%)
& 16/24 (66.67\%$\pm$11.79\%)
& 13/24 (54.17\%$\pm$11.79\%) \\

Pour Tea
& 13/24 (54.17\%$\pm$15.59\%)
& \textbf{18/24 (75.00\%$\pm$0.00\%)}
& 14/24 (58.33\%$\pm$5.89\%)
& 16/24 (66.67\%$\pm$15.59\%)
& 16/24 (66.67\%$\pm$15.59\%)
& 16/24 (66.67\%$\pm$5.89\%)
& 16/24 (66.67\%$\pm$5.89\%) \\

Lid on Teapot
& 15/24 (62.50\%$\pm$10.21\%)
& 11/24 (45.83\%$\pm$11.79\%)
& 10/24 (41.67\%$\pm$11.79\%)
& 18/24 (75.00\%$\pm$0.00\%)
& 16/24 (66.67\%$\pm$11.79\%)
& \textbf{20/24 (83.33\%$\pm$11.79\%)}
& 14/24 (58.33\%$\pm$15.59\%) \\

Stack Cup
& 8/24 (33.33\%$\pm$11.79\%)
& 14/24 (58.33\%$\pm$15.59\%)
& 9/24 (37.50\%$\pm$0.00\%)
& \textbf{20/24 (83.33\%$\pm$5.89\%)}
& \textbf{20/24 (83.33\%$\pm$5.89\%)}
& 16/24 (66.67\%$\pm$11.79\%)
& 16/24 (66.67\%$\pm$11.79\%) \\

\midrule
\multicolumn{8}{l}{\textit{Multi-stage}} \\

Plates on Stand
& 0/24 (0.00\%$\pm$0.00\%)
& 2/24 (8.33\%$\pm$5.89\%)
& 5/24 (20.83\%$\pm$11.79\%)
& \textbf{13/24 (54.17\%$\pm$11.79\%)}
& 11/24 (45.83\%$\pm$11.79\%)
& 10/24 (41.67\%$\pm$15.59\%)
& 7/24 (29.17\%$\pm$5.89\%) \\

Stack Cups
& 5/24 (20.83\%$\pm$5.89\%)
& 12/24 (50.00\%$\pm$10.21\%)
& 8/24 (33.33\%$\pm$15.59\%)
& \textbf{18/24 (75.00\%$\pm$0.00\%)}
& \textbf{18/24 (75.00\%$\pm$0.00\%)}
& 13/24 (54.17\%$\pm$15.59\%)
& 13/24 (54.17\%$\pm$15.59\%) \\

\midrule
Total
& 71/216 (32.87\%)
& 95/216 (43.98\%)
& 102/216 (47.22\%)
& 160/216 (74.07\%)
& 155/216 (71.76\%)
& 146/216 (67.59\%)
& 130/216 (60.19\%) \\

\bottomrule
\end{tabular}
}\vspace{-13pt}
\end{table*}

\subsection{Ablation on Sampling Constraints/Validation Program}
\label{sec:ablation}

We evaluate the contributions of the VLM-generated sampling constraints
$\mathcal{S}$ and validation program $\mathcal{V}$ on three representative tasks:
\texttt{Mug on Rack}, \texttt{Plate on Stand}, and
\texttt{Mug on Plate}. We compare the mesh-based full method against three variants.
\textbf{Default $\mathcal{V}$} replaces the VLM-generated validation program with a
task-agnostic default $\mathcal{V}$ that imposes no contact or XYZ-relation
constraints, uses the default penetration thresholds, and assigns moderate
position and orientation drift tolerances. \textbf{Default $\mathcal{S}$} retains
the VLM-generated $\mathcal{V}$ but replaces $\mathcal{S}$ with a maximally inclusive default
sampling distribution, using the largest translational and rotational
ranges on all six axes and no directional bias. \textbf{Default $\mathcal{S}+\mathcal{V}$} replaces both components with their corresponding defaults.

We report task success and $N_{\mathrm{consistent}}$ across 12 trials
with randomized object placements. Task success includes final VLM
candidate selection and execution, using a maximum sampling budget of
$N_s=2000$. Sampling stops once ten candidates are accepted.

\begin{figure}[t]
    \centering
    \begin{minipage}[t]{0.48\columnwidth}
        \centering
        \textbf{\small Task-consistent Candidates}
        
        \vspace{2mm}
        \includegraphics[width=\linewidth]{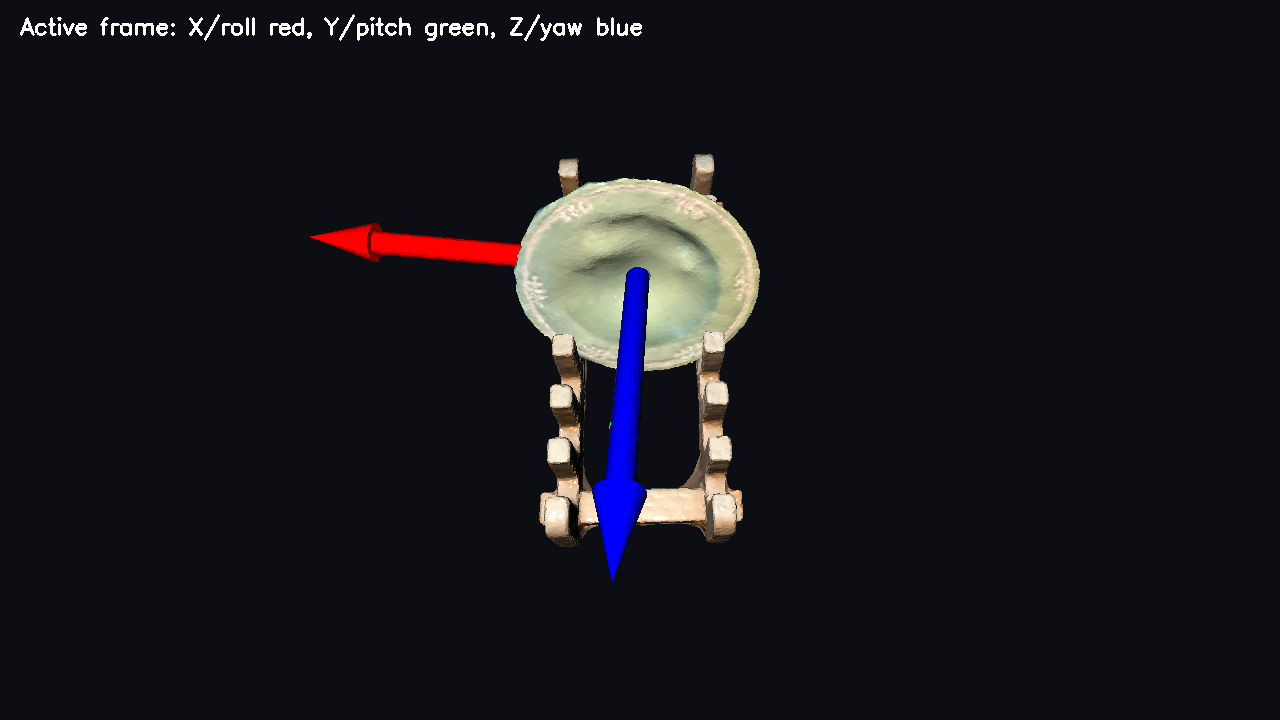}
        
        \vspace{2mm}
        \includegraphics[width=\linewidth]{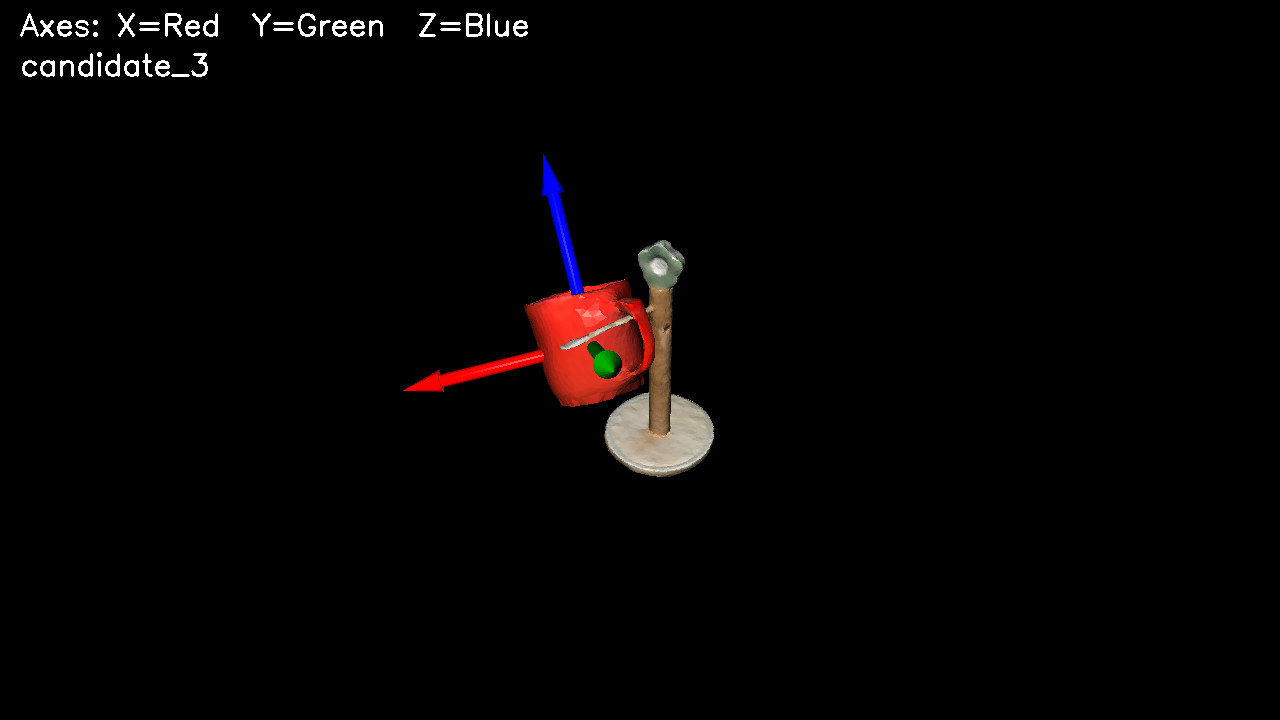}
    \end{minipage}
    \hfill
    \begin{minipage}[t]{0.48\columnwidth}
        \centering
        \textbf{\small Task-Inconsistent Candidates}
        
        \vspace{2mm}
        \includegraphics[width=\linewidth]{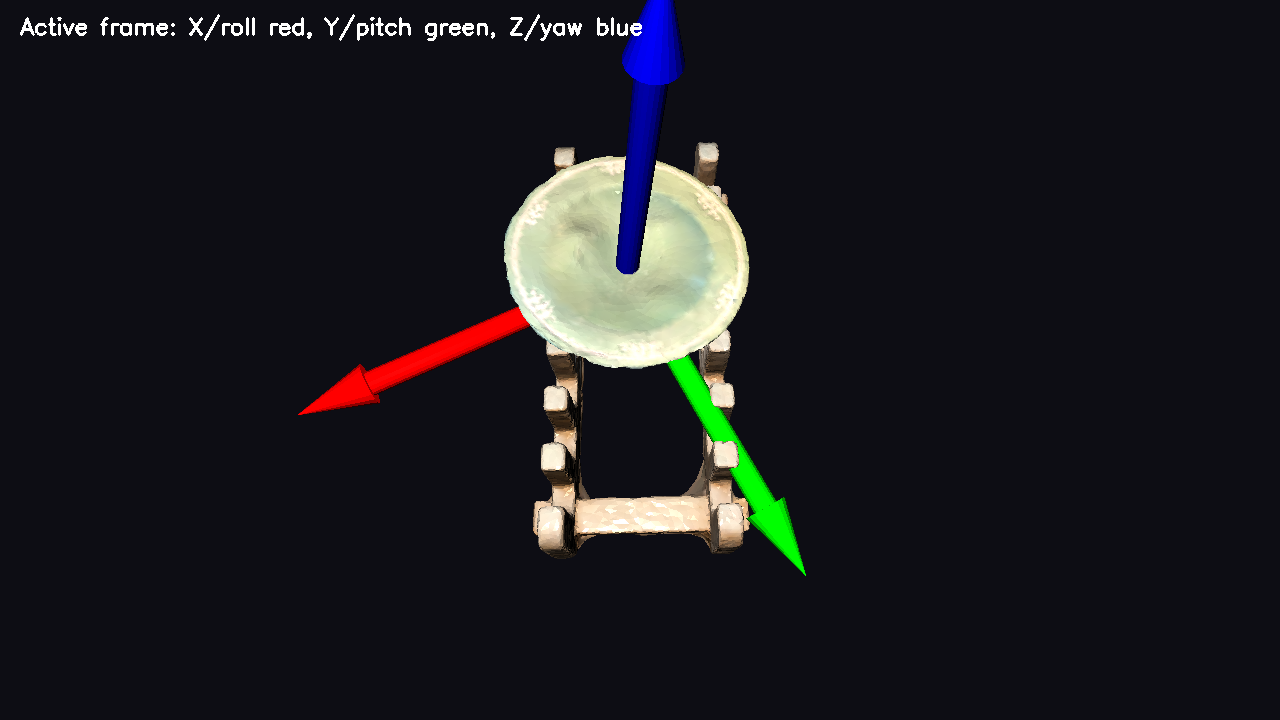}
        
        \vspace{2mm}
        \includegraphics[width=\linewidth]{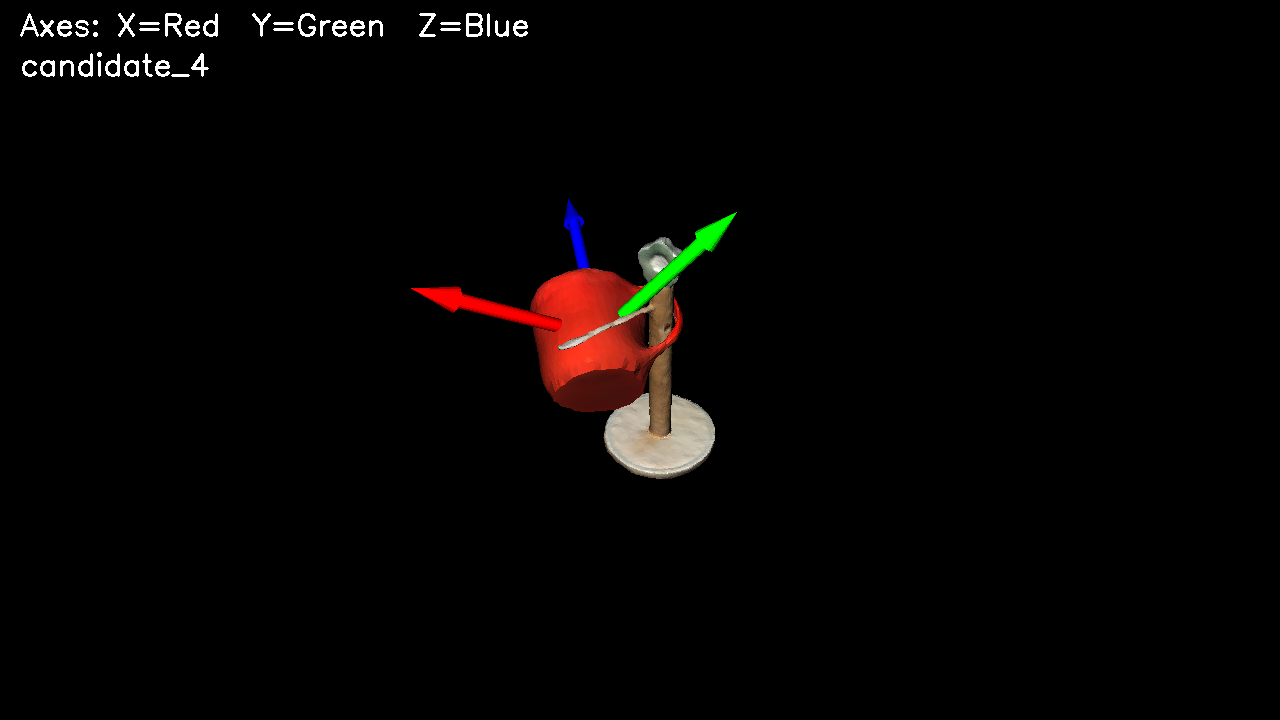}
    \end{minipage}

    \caption{Task-consistent and task-inconsistent candidates.}\vspace{-10pt}
    \label{fig:false-positives}
\end{figure}

$N_{\mathrm{consistent}}$ counts task-consistent poses among the first
ten candidates accepted by each variant's validation program, averaged
across trials. We report it at budgets
$N_s\in\{250,500,1000,2000\}$, counting missing candidates as inconsistent.
This metric captures both accepted-candidate quality and sampling
efficiency. As illustrated in Fig.~\ref{fig:false-positives}, a candidate may satisfy $\mathcal{V}$ while
failing to realize the intended arrangement. False-positive candidates
can therefore confuse final VLM selection even when the candidate set
contains a task-consistent pose.

As Table~\ref{tab:ablation} shows, across all three tasks, the generated $\mathcal{S}$ yields task-consistent
candidates at smaller sampling budgets by concentrating proposals in
task-relevant regions, while the generated $\mathcal{V}$ improves candidate quality
further by setting physical criteria suitable for the task. Consequently, the full method
consistently attains the highest $N_{\mathrm{consistent}}$ and saturates
by $N_s=500$ for \texttt{Mug on Plate} and by $N_s=1000$ for \texttt{Plate on Stand}; 
\texttt{Mug on Rack} continues to improve at larger budgets. Replacing $\mathcal{V}$ with its task-agnostic default lowers $N_{\mathrm{consistent}}$, yielding more candidates that are physically plausible but do not realize the instruction. Removing both $\mathcal{S}$ and $\mathcal{V}$ combines broad sampling with weak task-specific filtering and performs worst overall.

\begin{table}[t]
\centering
\caption{Ablation of sampling constraints $\mathcal{S}$ + validation program $\mathcal{V}$.}
\label{tab:ablation}
\resizebox{\columnwidth}{!}{
\begin{tabular}{llccccc}
\toprule
\textbf{Task} & \textbf{Variant}
& \textbf{Success Rate}
& \multicolumn{4}{c}{$\mathbf{N_{\mathrm{consistent}}}$} \\
\cmidrule(lr){4-7}
& & 
& $N_s=250$
& $N_s=500$
& $N_s=1000$
& $N_s=2000$ \\
\midrule

\multirow{4}{*}{Mug on Rack}
& Full method ($\mathcal{S}+\mathcal{V}$)
& \textbf{7/12}
& \textbf{3.17} & \textbf{4.50} & \textbf{4.92} & \textbf{5.50} \\
& Default $\mathcal{V}$
& 6/12
& 0.92 & 2.08 & 2.67 & 3.08 \\
& Default $\mathcal{S}$
& 4/12
& 0.42 & 0.75 & 1.50 & 2.17 \\
& Default $\mathcal{S}+\mathcal{V}$
& 2/12
& 0.33 & 0.42 & 0.67 & 0.75 \\
\midrule

\multirow{4}{*}{Plate on Stand}
& Full method ($\mathcal{S}+\mathcal{V}$)
& \textbf{8/12}
& \textbf{5.67} & \textbf{7.17} & \textbf{7.92} & \textbf{7.92} \\
& Default $\mathcal{V}$
& 5/12
& 3.67 & 4.00 & 4.08 & 4.08 \\
& Default $\mathcal{S}$
& 6/12
& 2.58 & 3.17 & 3.67 &  3.92 \\
& Default $\mathcal{S}+\mathcal{V}$
& 2/12
& 0.58 & 1.17 & 1.25 & 1.25 \\
\midrule

\multirow{4}{*}{Mug on Plate}
& Full method ($\mathcal{S}+\mathcal{V}$)
& \textbf{11/12}
& \textbf{7.25} & \textbf{7.58} & \textbf{7.58} & \textbf{7.58} \\
& Default $\mathcal{V}$
& 5/12
& 2.75 & 2.75 & 2.75 & 2.75 \\
& Default $\mathcal{S}$
& 7/12
& 3.17 & 3.75 & 3.83 & 3.83 \\
& Default $\mathcal{S}+\mathcal{V}$
& 3/12
& 1.58 & 1.58 & 1.58 & 1.58 \\

\bottomrule
\end{tabular}
}
\end{table}

\subsection{Ablation on Entire Pose Refinement Process}
\begin{figure}[t]
    \centering
    \includegraphics[width=\columnwidth]{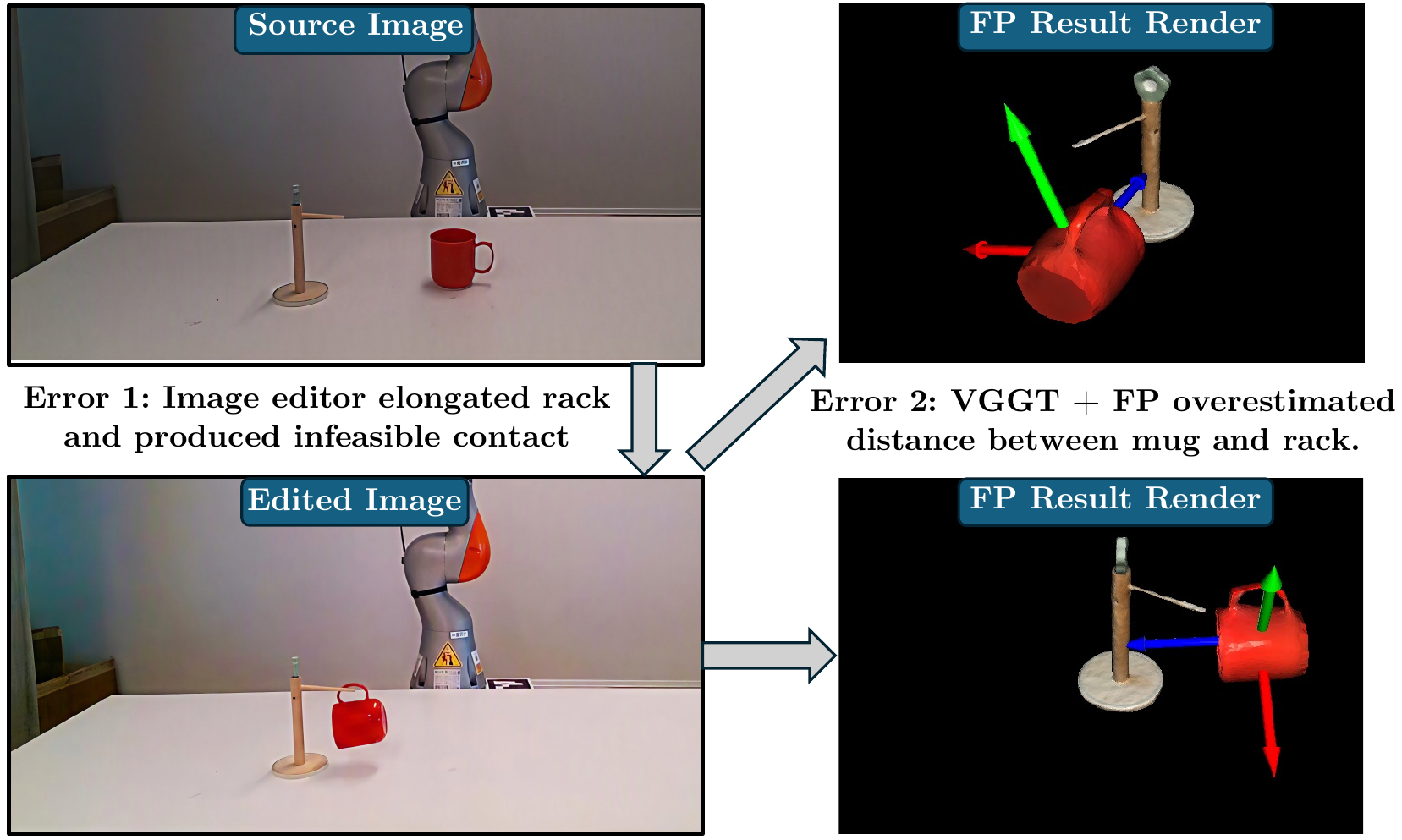}
    \caption{Example showing sources of error that could arise when extracting a goal pose from the VLM-edited image. First, the image editor elongates the rack's branch and produces a physically infeasible contact arrangement. Second, the perception module with VGGT and FoundationPose overestimates the distance between the mug and the rack.\vspace{-15pt}}
    \label{fig:error-sources}
\end{figure}

To assess the necessity of simulation-based validation in VerifLM, we evaluate whether the initial pose recovered directly from the edited image is sufficient to serve as the final manipulation goal. In this ablation, we terminate the pipeline immediately after relative goal pose recovery (Sec. \ref{sec:relative-goal-pose-recovery}), and use the resulting image-derived pose as the target, removing VLM-generated sampling constraints, physical validation, and candidate selection. This ablation uses predefined object meshes and reports task-planning
success, isolating pose refinement from object-model estimation. Fig.~\ref{fig:error-sources} illustrates why this direct approach can fail. Errors can arise both from the edited image itself and from the subsequent perception pipeline. In the example, the image editor elongates the rack branch and depicts a physically infeasible mug--rack contact, while depth and pose estimation further overestimate the separation between the two objects.

\begin{table}[t]
\centering
\caption{Ablation of the pose-refinement process.}
\label{tab:pose_refinement_ablation}
\resizebox{\columnwidth}{!}{
\begin{tabular}{lcc}
\toprule
\textbf{Task} & \textbf{Image-Derived Pose Only} & \textbf{Full Method} \\
\midrule
Mug on Rack      & 10/24  & \textbf{16/24} \\
Mug on Plate     & 19/24 & \textbf{22/24} \\
Mug on Stand     & 11/24 & \textbf{20/24} \\
Plate on Stand   & 5/24  & \textbf{17/24} \\
Pour Tea         & 14/24 & \textbf{16/24} \\
Lid on Teapot    & 16/24 & \textbf{18/24} \\
Stack Cup        & 6/24  & \textbf{20/24} \\
Plates on Stand  & 0/24  & \textbf{13/24} \\
Stack Cups       & 3/24  & \textbf{18/24} \\
\midrule
\textbf{Total}   & 84/216 & \textbf{160/216} \\
\bottomrule
\end{tabular}}\vspace{-10pt}
\end{table}

Both variants are evaluated in the same predefined-mesh digital twin using the task-planning success criteria described in Sec.~\ref{sec:quantitative-results}. Table~\ref{tab:pose_refinement_ablation} compares the image-derived pose against the full pose-refinement pipeline. 
Directly using the image-derived pose succeeds in only 84/216 trials (38.89\%), whereas the full method succeeds in
160/216 trials (74.07\%). This gap reflects errors from both image editing and subsequent pose estimation, which can place the recovered
pose near the desired configuration yet still require correction. Even when the generated image appears task-consistent and physically
plausible, there is no guarantee that a valid pose can be recovered from it, as shown in Fig.~\ref{fig:error-sources}. These results reinforce that the edited
image provides a useful initial task specification, but its recovered pose cannot be trusted without simulation-based verification, which VerifLM provides.

\subsection{Ablation on Parallel-World Validation}
\label{sec:parallel-world-ablation}

Finally, in mesh-free-input mode, we evaluate whether parallel-world validation improves execution success over validation in a single estimated world. Both variants use the same reconstructed meshes, sampling constraints, validation program, sampling budget of $N_s=200$, and sampled poses. \textit{Single World} uses a mesh scale of 1.0 and the first VLM-predicted physical parameters, retaining the first five valid candidates. \textit{Parallel Worlds} evaluates each candidate across the 27 geometry--physics worlds described in Sec.~\ref{sec:mujoco-sample-and-validation} and retains the five candidates with the highest robustness scores. In both variants, the VLM selects the final goal from the retained candidates.

\begin{table}[t]
\centering
\caption{Ablation of parallel-world validation.}
\label{tab:parallel_world_ablation}
\resizebox{\columnwidth}{!}{
\begin{tabular}{lcc}
\toprule
\textbf{Task}
& \textbf{Single-World Validation}
& \textbf{Parallel-World Validation} \\
\midrule
Mug on Rack      & 11/24 & \textbf{14/24} \\
Mug on Plate     & 19/24 & \textbf{20/24} \\
Mug on Stand     & 16/24 & \textbf{17/24} \\
Plate on Stand   & 9/24 & \textbf{13/24} \\
Pour Tea         & 15/24 & \textbf{16/24} \\
Lid on Teapot    & 7/24 & \textbf{14/24} \\
Stack Cup        & 9/24 & \textbf{16/24} \\
Plates on Stand  & 2/24 & \textbf{7/24} \\
Stack Cups       & 4/24 & \textbf{13/24} \\
\midrule
\textbf{Total}
& 92/216
& \textbf{130/216} \\
\bottomrule
\end{tabular}}\vspace{-13pt}
\end{table}

Table~\ref{tab:parallel_world_ablation} shows that parallel-world validation improves execution success on all nine tasks, increasing the aggregate success rate from 92/216 (42.6\%) to 130/216 (60.2\%). Single-world validation evaluates candidates only against one pair of nominal meshes and is therefore subject to geometric inaccuracies in estimated meshes. Consequently, it can retain poses that satisfy the validation program for the reconstructed objects but fail to realize the required contacts or support on the true objects. Single-world validation can also reject feasible goals when, for example, an overestimated active-cup size or underestimated passive-cup opening makes insertion infeasible in the nominal model, even when the real cups fit. Parallel-world validation instead evaluates each candidate across variations in object scales and estimated physical parameters, favoring poses that remain valid under model uncertainty. Its higher execution success indicates that robustness across plausible digital twins transfers more reliably to the real objects.

\section{Conclusion}
We presented VerifLM, a framework for refining image-derived manipulation goals into physically reliable 6-DoF object poses using VLM-generated sampling constraints and task-specific digital-twin validation. Its mesh-free-input variant reconstructs object geometry, estimates physical-parameter hypotheses, and evaluates candidates across parallel worlds to favor goals robust to pose, scale, and physical uncertainty. However, its performance still depends on reconstruction quality, since missing or distorted contact features can prevent the digital twins from accurately modeling the task. The method also inherits variability from VLM-generated edits and reasoning, which may lead to inconsistent sampling constraints or validation programs. Future work will address these limitations through improved uncertainty modeling that accounts for geometric feature distortions and explore ways to improve robustness to VLM hallucinations.

\bibliographystyle{IEEEtran}
\bibliography{references}

\end{document}